\documentclass[runningheads]{llncs}
\usepackage[T1]{fontenc}
\usepackage[table]{xcolor}
\usepackage{tabularx}
\usepackage{float}
\usepackage{amsmath,amssymb}
\usepackage{subcaption}
\usepackage{arydshln}
\usepackage{multirow}
\usepackage[ruled,vlined]{algorithm2e}
\usepackage{pifont}
\usepackage{graphicx}

\newcommand{\ie}{\textit{i.e.}}
\newcommand{\eg}{\textit{e.g.}}

\newcommand{\cmark}{\ding{51}}%
\newcommand{\xmark}{\ding{55}}%

\newcommand{\shorthypen}{\scalebox{0.8}[1]{-}}

\begin{document}

\title{Federated Medical Image Classification\\ under Class and Domain Imbalance\\ exploiting Synthetic Sample Generation}
\titlerunning{Federated Unbalanced Classification via Generative Sampling}

\author{Martina Pavan\inst{1, 2}\orcidID{0009\shorthypen0005\shorthypen4847\shorthypen3004} \and
Matteo Caligiuri\inst{1}\orcidID{0009\shorthypen0006\shorthypen2928\shorthypen1047} \and
\mbox{Francesco Barbato\inst{1}\orcidID{0000\shorthypen0001\shorthypen9893\shorthypen5813} \and
Pietro Zanuttigh\inst{1}\orcidID{0000\shorthypen0002\shorthypen9502\shorthypen2389}}}

\institute{University of Padova, Padova (PD), Italy \\
\email{\{martina.pavan,matteo.caligiuri,francesco.barbato,zanuttigh\}@dei.unipd.it} 
\and NIDEK Technologies Srl, Albignasego (PD), Italy \\
\email{martinapavan@nidektechnologies.it}
}
\maketitle              %
\begin{abstract}
Exploiting deep learning in medical imaging faces critical challenges, including strict privacy constraints, heterogeneous imaging devices with varying acquisition properties, and class imbalance due to the uneven prevalence of pathologies. In this work, we propose FedSSG, a novel Federated Learning framework that addresses domain shifts caused by diverse imaging devices while mitigating the under-representation of rare pathologies. The key contribution is a strategy for generating synthetic samples and distributing them across clients to improve coverage of both underrepresented pathologies and imaging devices. Experimental results demonstrate that our approach significantly enhances model performance and generalization across heterogeneous institutions, with minimal computational overhead at the client side.

\end{abstract}

\section{Introduction} \label{sec:intro}
Deep learning has become a key tool in medical image analysis, enabling significant progress in tasks such as disease detection, segmentation, and classification. However, its adoption is limited by strict privacy regulations, \eg, the General Data Protection Regulation (GDPR), and the sensitive nature of patient data, which prevents the exchange of medical images across institutions~\cite{kaissis2020secure}. Federated Learning (FL) offers a solution by allowing multiple institutions to collaboratively train models without exchanging local data~\cite{mcmahan2017communication,kairouz2021advances}. In FL, each client trains a local model on its private data and shares only model updates with a central server, thus preserving privacy while leveraging a larger effective dataset.

A major challenge when applying FL in the medical imaging field is data heterogeneity. Images collected at different sites have different appearances due to differences in imaging devices, acquisition protocols, patient populations, and annotation practices~\cite{li2020federated,zhou2021fed}. 
Variations in scanner models, resolution, or contrast settings introduce domain shifts, while differences in demographics or disease prevalence lead to uneven distributions of pathologies across clients. Such non-IID %
data can degrade federated model performance~\cite{sheller2020federated,kaissis2021secure}.

Class imbalance is another critical issue: common diseases dominate hospital records, while rare conditions are under-represented~\cite{litjens2017survey,shen2017deep}. 
In FL settings, this imbalance is often exacerbated, as each client may observe only a limited subset of classes, leading to biased local updates and reduced model generalization.

In this work, we focus on a realistic classification scenario -- using skin lesion classification as a benchmark due to the availability of public data -- where each client in a federated setup represents a different institution, and images are acquired using diverse types of machinery (\eg, dermatoscopes). This introduces pronounced domain heterogeneity, as each client’s data distribution is strongly influenced by the specific imaging device used. 
In addition, both class imbalance (due to the rarity of pathologies) and domain imbalance (due to device utilization and costs) are present, further biasing training and limiting generalization.

To address these challenges, we propose a novel FL framework that leverages a public dataset to pretrain a global model, providing a robust and stable initialization across clients. 
Building on this initialization, a class-conditional generative model is employed to mitigate the under-representation of both pathologies and imaging devices, thereby enhancing data coverage during federated training. 
By explicitly addressing class and domain imbalance within the federated learning process, our framework promotes domain generalization and robust performance across heterogeneous domains while preserving data privacy.

\section{Related Works} \label{sec:related}
\noindent \textbf{Federated Learning (FL)} is a machine learning approach designed to handle decentralized learning (distributed optimization) without sharing private data. 
Nowadays, due to rising privacy and legal concerns, FL has become crucial for many vision tasks~\cite{shenaj2023federated}.
The FL framework was introduced in \cite{mcmahan2017communication}, which proposes performing a weighted average of the local (client-side) models after each distributed optimization round, before propagating the aggregated model to the clients as initialization for the subsequent step. 
The approach (FedAvg) is very effective in simple scenarios where there are few clients and the samples are well-distributed among them; however, it struggles with non-IID data distributions.
A later improvement is proposed in FedProx~\cite{li2020federated}, which adds a proximal term to the local objectives to limit the impact of local updates, reduce client drift, and achieve a more stable training evolution.
In a similar vein are SCAFFOLD~\cite{karimireddy2020scaffold}, which estimates client drift by comparing server and client update directions and using it for correction, and MOON~\cite{li2021model}, which improves local training in non-IID settings through model-based contrastive learning, enhancing novel client knowledge while preserving global server knowledge.
Other works~\cite{zhuang2025foundationmodelmeetsfederated,li2024synergizing,caligiuri2025fedpromo} combine federated learning with foundation models in various ways, e.g., using them to stabilize the training or exploiting their well-structured latent space.

\noindent \textbf{Pretraining in FL} has recently gained traction, as it allows a designer to initialize the client models to a state that better suits federated training than random initialization.
More specifically, traditional FL settings~\cite{mcmahan2017communication} limit the role of the server solely to that of model aggregator and fleet controller. 
However, given its computational capabilities, which often overshadow those of the clients, it is advantageous to exploit it for improved efficiency and performance. 
Therefore, recent works have begun exploring pretraining at server-side before initiating local training on devices~\cite{shenaj2023federated}.
Pretraining addresses data heterogeneity, achieves longer local training, and reduces communication costs \cite{nguyen2023where,chen2023on}.
Beyond communication efficiency, pretraining enables more realistic FL scenarios by addressing tasks where client-side annotations are impractical.
For example, pretraining on synthetically generated supervised data facilitates unsupervised client-side learning for challenging vision tasks such as semantic segmentation \cite{shenaj2023learning}.

\noindent \textbf{FL in Medical Imaging} has seen growing adoption due to the sensitive nature of patient data and strict privacy regulations~\cite{guan2024federated,nazir2023diagnostics}, but it still faces unique challenges. 
Data distributions are often highly heterogeneous across clients due to variations in patient populations, acquisition devices, and acquisition protocols. 
In dermatoscopic image classification, for example, clients may have different distributions of lesion types, and images may have been captured using different devices, introducing substantial domain shifts. 
These non-IID characteristics can degrade the performance of standard federated algorithms~\cite{zhou2025flmedical}.
Class imbalance is another critical problem. Certain lesion types are rare and underrepresented in local datasets, leading to biased models if not addressed. Techniques such as FedIIC~\cite{wu2023fediic} propose tailored representation learning and classifier adjustments to mitigate this imbalance in FL settings, since asynchronous update strategies are effective for skin lesion classification under limited and skewed datasets, allowing the model to adapt more robustly to heterogeneous client data.
Generative models offer a promising complementary approach. By synthesizing additional images, these models can balance both class distributions and acquisition domains locally on each client without sharing sensitive data. For instance, FedGAN~\cite{kamran2025fedgan} demonstrates how federated generative models can produce realistic medical images for tasks such as diabetic retinopathy detection, improving data coverage, and model generalization.
Building on these insights, our work leverages a generative model to augment dermatoscopic images for each client, simultaneously addressing both class and device imbalance. 
This strategy allows the federated model to learn more robust and generalizable features, thereby improving classification performance while respecting privacy constraints.

\section{Problem Formulation} \label{sec:problem}
We consider a classification task on a private dataset \mbox{$D_{\text{priv}} \!=\! \{(\mathbf{x}^{(i)}_{\text{priv}}, y^{(i)}_{\text{priv}})\}_{i=1}^{n_{\text{priv}}}$}, where $\mathbf{x} \in \mathcal{X} \subset \mathbb{R}^{H\times W\times3}$ represents an input image and
$y \in \mathcal{Y} = \{1,\dots,C\}$ denotes the corresponding class label, among a set of $C$ classes.
Due to privacy constraints, samples in $D_{\text{priv}}$ cannot be shared across institutions. The private dataset is composed of disjoint subsets acquired using different devices. Let \mbox{$\mathcal{J} = \{1,\dots,J\}$} denote the set of device domains, such that $D_{\text{priv}} = \bigcup_{j \in \mathcal{J}} D^j_{\text{priv}}$, where each $D^j_{\text{priv}}$ is drawn from a device-specific data distribution
$P^j_{\text{priv}}(\mathbf{x}, y)$.
For different devices $i \neq j$, we generally have
$P^i_{\text{priv}}(\mathbf{x}, y) \neq P^j_{\text{priv}}(\mathbf{x}, y)$,
resulting in domain heterogeneity across  private data. In addition, we assume access to a public dataset
\mbox{$D_{\text{pub}} = \{(\mathbf{x}^{(i)}_{\text{pub}},  {y}^{(i)}_{\text{pub}})\}_{i=1}^{n_{\text{pub}}}$} whose data distribution is related to, but does not match, that of the private one. The public data can be freely accessed and shared, but does not contain information about acquisition devices.

Our goal is to learn a global classification model $\mathcal{M}_\theta$, parameterized by $\theta$, that effectively leverages both datasets through a federated learning framework, thus preserving the privacy of $D_{\text{priv}}$. 
In the federated learning scenario, the private dataset is scattered among $K$ clients $\{k\}_{k=1}^K$, where each client $k$ holds a local dataset $D_k \subseteq D^{d(k)}_{\text{priv}}$ drawn from a single device domain $d(k) \in \mathcal{J}$. %
Each client trains a local model on its own data and communicates model updates to a central server, which aggregates the updates and redistributes the global model to all clients. This iterative process continues for $R$ communication rounds.
Following realistic medical imaging scenarios, both the public and private datasets exhibit class ($P(y)$ is non-uniform) and domain %
imbalance.
As a result, clients associated with smaller domains and rarer classes contribute fewer samples during training, leading to biased optimization.

Overall, the learning problem is characterized by three intertwined challenges:
(1) statistical heterogeneity across device domains,
(2) class imbalance within local client datasets, and
(3) unequal representation of domains across the federated system.
These factors jointly hinder the convergence and generalization of standard federated learning methods. In the following, we introduce a domain-aware federated learning framework designed to address these challenges.

\section{Proposed Method} \label{sec:method}
\begin{figure*}[h]
    \centering
    \includegraphics[width=\textwidth]{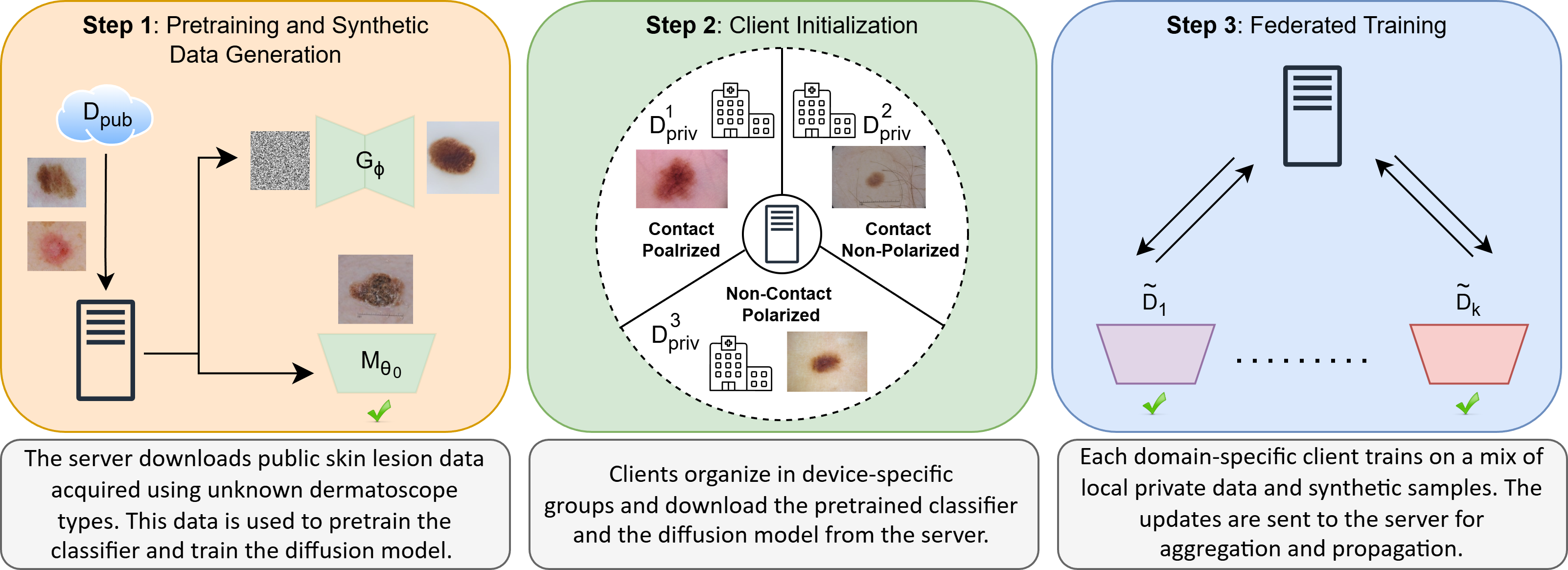}
    \caption{Architecture of our federated multi-domain classification approach (FedSSG). On the left, the server-side pretraining, where the classifier and the diffusion model are trained on public mixed-sensor data. In the middle, the per-sensor organization of the clients, and federated training initialization. On the right, a schematic view of local training and federated aggregation.}
    \label{fig:arch}
\end{figure*}

The proposed framework (FedSSG) addresses the challenges of domain heterogeneity and data imbalance in federated learning by integrating three key components: (1) pretraining on public data, (2) domain-aware generalization, and (3) class- and domain-conditional data augmentation. 
The overall pipeline begins by pretraining a global model $\mathcal{M}_{\theta_0}$  on the public dataset $D_{\text{pub}}$, providing a robust initialization $\theta_0$ shared across all clients. 
Clients are then grouped according to the corresponding acquisition device, forming domain-specific clusters $\{\mathcal{C}_j\}_{j=1}^J$ that reflect the underlying distributional differences across imaging domains. 
To alleviate class and domain imbalance, each client leverages additional samples from the generative model $\mathcal{G}_\phi$  for underrepresented classes, thereby improving the representativeness of local data. 
Finally, the adapted client models are iteratively trained and aggregated over $R$ communication rounds, resulting in the final global model $\mathcal{M}_\theta$. 
In Figure \ref{fig:arch}, we show a schematic breakdown of the proposed pipeline.

\subsection{Pretraining on Public Data } \label{sub:method:pretrain}

To provide a strong initialization for the federated training phase, the global model is first pretrained on the public dataset $D_{\text{pub}}$, obtaining an initial model $\mathcal{M}_{\theta_0}$.
The objective of this stage is to learn generalized visual representations relevant to the target pathology classification task, prior to exposure to heterogeneous private data.
More in detail, the model parameters are optimized by minimizing the cross-entropy loss on $D_{\text{pub}}$, \ie, $\theta_0 = \arg\min_{\theta} \mathcal{L}_{\text{pub}}(\theta)$, where $\mathcal{L}_{\text{pub}}$ is the empirical cross-entropy computed over $D_{\text{pub}}$.
The resulting parameters $\theta_0$ serve as a domain-agnostic initialization for federated training, enabling faster convergence and improved generalization across heterogeneous domains.

\subsection{Data Augmentation with Synthetic Samples} \label{sub:method:augmentation}

Both public and private datasets exhibit class imbalance. Moreover, the private dataset suffers from domain imbalance across acquisition devices, as the sizes of device-specific subsets vary significantly between domains.
These factors bias the empirical risk minimization objective, causing common classes and larger domains to dominate the training and limit generalization across different clients.

\subsubsection{Synthetic Sample Generation}
To mitigate these imbalances, we adopt a class-conditional generative model
$\mathcal{G}_\phi : \mathcal{Z} \times \mathcal{Y} \rightarrow \mathcal{X}$
that approximates the conditional distribution $P(\mathbf{x} \! \mid \! y)$.
The generator $\mathcal{G}_\phi$ is implemented as a denoising diffusion model with a U-Net backbone, incorporating ConvNeXt-style~\cite{liu2022convnet2020s} residual blocks, multi-resolution attention layers, and FiLM-based~\cite{perez2017filmvisualreasoninggeneral} conditioning.
Temporal information is injected through sinusoidal timestep embeddings, while class information is encoded via learnable class embeddings.

The generator is trained %
by the server on the public dataset $D_{\text{pub}}$ prior to federated learning (FL) deployment.
This design prevents leakage of private client statistics while providing a shared semantic prior that can be leveraged by all clients.
To improve robustness and training stability, class conditioning is randomly dropped during training, following a classifier-free guidance strategy.

During FL, the generator remains frozen and is used to produce synthetic samples
$\tilde{\mathbf{x}} = \mathcal{G}_\phi(\mathbf{z}, y)$,
where $\mathbf{z}$ is sampled from a predefined latent prior.
Sample generation follows an iterative denoising process with classifier-free guidance, after which the generated images are upscaled using a pretrained super-resolution model to preserve high-frequency details. Several examples of generated samples are in the \textit{Supplementary Material}.

\subsubsection{Synthetic Sample Allocation} To limit class and domain imbalance, we introduce a custom sample distribution that allocates proportionally more generated data to data-sparse clients as follows:
let each client $k \in \{1,\dots,K\}$ hold a local dataset
$D_k = \{(\mathbf{x}_k^{(i)}, y_k^{(i)})\}_{i=1}^{n_k}$,
for each class $c \in \mathcal{Y}$ at client $k$, we define a class imbalance weight:
\begin{equation}
    w_{k,c} = \max_{c' \in \mathcal{Y}} n_{k,c'} - n_{k,c} + \epsilon,
\end{equation}
where $n_{k,c}$ denotes the number of samples of class $c$ at client $k$, and $\epsilon > 0$ ensures that every class receives at least a small number of synthetic samples.

Let $d(k) \in \mathcal{J}$ be the domain of client $k$, and $N_{d(k)} = \sum_{i \in \mathcal{C}_{d(k)}} |D_i|$ be the total number of real samples available in domain $d(k)$, where $\mathcal{C}_{d}$ denotes the set of clients belonging to domain $d$.
We introduce a domain-dependent scaling factor $S_{d(k)}$ empirically tuned following the rule that $N_{d(i)} > N_{d(j)}$ implies $S_{d(i)} < S_{d(j)}$ to ensure that rarer domains appear more frequently  %
(for more details see the ablation in Table~\ref{tab:gen_num} that reports the quantitative results achieved with different settings of $S_{d(k)}$). %
Given this, the number of generated synthetic samples  for class $c$ at client $k$ is:
\begin{equation}
    n_{k,c}^{\text{synth}} =
    S_{d(k)} \cdot
    \frac{w_{k,c}}{\sum_{c' \in \mathcal{Y}} w_{k,c'}}.
\end{equation}
Finally, %
the augmented dataset for client $k$ is $\tilde{D}_k = D_k \cup \bigcup_{c \in \mathcal{Y}} \left\{(\tilde{\mathbf{x}}_{k,c}^{(i)}, c)\right\}_{i=1}^{n_{k,c}^{\text{synth}}}$.

\subsubsection{Privacy Considerations} The proposed allocation strategy relies solely on aggregate statistics available on the client-side (total number of local samples and their per-class counts). In our setting, these statistics do not pose a privacy risk, as they do not contain patient-level information and are independent of individual data samples. No raw data, intermediate representations, or private domain-specific features are shared across clients or with the central server.

\subsection{Federated Training with Synthetic Samples}
Finally, the model $\mathcal{M}_\theta$ is trained in a federated fashion.
At each round, each client $k$ performs a local training procedure using the augmented dataset $\tilde{D}_k$ and then sends the updated model weights to the server, which aggregates them using standard federated averaging (FedAvg)~\cite{mcmahan2017communication}.

The core idea is to reduce the effects of class and domain imbalance that affect the training data by acting directly on the edge devices.
During the federated rounds, each client exploits the pretrained generative model to augment its local training data, alleviating bias induced by locally overrepresented classes.
This leads to more stable client updates, which in turn improve the accuracy and generalization of the global model.

\section{Experimental Setup} \label{sec:experiments}
\subsection{Implementation Details} \label{sub:experiments:implementation}
\subsubsection{Classifier Architecture}
Following common practice in various medical image analysis works, %
we adopt EfficientNet-B0 \cite{tan2019efficientnet} as the backbone architecture for the classification model. The encoder $E$ is initialized with ImageNet-1k pretrained weights~\cite{deng2009imagenet} and is used for both the public data pretraining stage and the federated learning setup.
The classification head $H$ is implemented as a multilayer perceptron composed of two fully connected layers with 512 hidden units, each followed by batch normalization, ReLU activation, and dropout with probability 0.3. A final linear layer maps the learned features to the $C$ output classes.
Input images are resized to $224 \times 224$ pixels to match the pretraining resolution. %

\subsubsection{Public Data Pretraining}
Data augmentation is applied only to underrepresented classes using the \textit{Albumentations}~\cite{info11020125} library. It includes random horizontal and vertical flips, $90^\circ$ rotations, and mild color perturbations (brightness, contrast, and hue). Mini-batches are sampled according to class-dependent probabilities inversely proportional to class frequency,
\[
p_i = \frac{1 / n_i}{\sum_{j=1}^{C} 1 / n_j},
\]
ensuring approximately class-balanced batches during training.
The model is optimized using AdamW~\cite{loshchilov2017decoupled} with learning rates $10^{-3}$ for $H$ and $10^{-4}$ for $E$, weight decay $10^{-4}$, and a ReduceLROnPlateau scheduler~\cite{loshchilov2017sgdr}.  
Training lasts 30 epochs with early stopping and batch size 32.

\subsubsection{Federated Setup}
We simulate federated deployment using the Flower~\cite{beutel2020flower} PyTorch library. For the main results (Sec. \ref{sub:experiments:main_results}), we consider a scenario with 85 total clients and 6 active clients per round. Training runs for 150 rounds. In Sec. \ref{sub:experiments:fed_config} we also show results under different configurations. Each active client trains its local model for 5 epochs per round, with batch size 32. Local optimization is performed using the AdamW optimizer~\cite{loshchilov2017decoupled} with $lr=10^{-4}$ and no weight decay.  

\subsubsection{Synthetic Data Generation}
The generator $\mathcal{G}_\phi$ is trained offline using only the public dataset $D_{\text{pub}}$.
All images are resized to $64\times64$ pixels and normalized to the range $[-1,1]$.
Training is performed using the Adam optimizer with a learning rate of $10^{-3}$ and a batch size of 128.
Gradient clipping with a maximum norm of 1.0 is applied for stability.
The diffusion model is trained for 50 epochs using 512 diffusion steps and cosine beta scheduler.
The training objective is the sum of the mean squared error (MSE) and the $L_1$ loss between the predicted and ground-truth noise.
Class conditioning is randomly dropped with probability $p = 0.1$ during training. For inference, we use classifier-free guidance with scale 5.0, and we upscale the resulting images to a spatial resolution of $256\times256$ using a pretrained Real-ESRGAN model~\cite{wang2021realesrgan}.
The synthetic samples are generated as detailed in Section \ref{sub:method:augmentation} using scaling factors $S_{d(k)} = (20, 50, 80)$ for CP, CNP, and NCP, respectively.

\newcommand{\gridimg}[1]{%
  \includegraphics[
    width=1\textwidth,
    height=1\textwidth,  %
    keepaspectratio=false,   %
    clip
  ]{#1}
}

\begin{figure*}[t]
    \centering

    \begin{subfigure}{\textwidth}
        \centering
        \begin{subfigure}[t]{.17\textwidth}
            \centering
            \gridimg{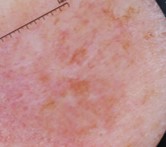}
        \end{subfigure}
        \begin{subfigure}[t]{.17\textwidth}
            \centering
            \gridimg{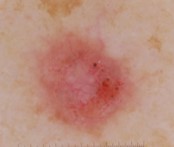}
        \end{subfigure}
        \begin{subfigure}[t]{.17\textwidth}
            \centering
            \gridimg{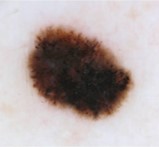}
        \end{subfigure}
        \begin{subfigure}[t]{.17\textwidth}
            \centering
            \gridimg{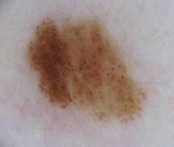}
        \end{subfigure}
        \begin{subfigure}[t]{.17\textwidth}
            \centering
            \gridimg{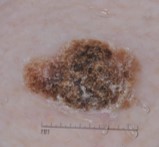}
        \end{subfigure}
    \end{subfigure}
    \begin{subfigure}{\textwidth}
        \centering
        \begin{subfigure}[t]{.17\textwidth}
            \centering
            \gridimg{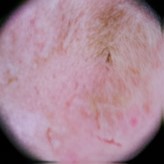}
            \parbox{\textwidth}{\centering\vspace{0.3em} Actinic \mbox{Keratosis}}
        \end{subfigure}
        \begin{subfigure}[t]{.17\textwidth}
            \centering
            \gridimg{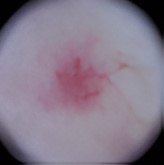}
            \parbox{\textwidth}{\centering Carcinoma}
        \end{subfigure}
        \begin{subfigure}[t]{.17\textwidth}
            \centering
            \gridimg{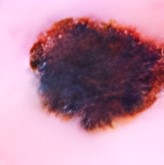}
            \parbox{\textwidth}{\centering Melanoma}
        \end{subfigure}
        \begin{subfigure}[t]{.17\textwidth}
            \centering
            \gridimg{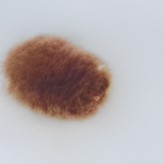}
            \parbox{\textwidth}{\centering Nevus}
        \end{subfigure}
        \begin{subfigure}[t]{.17\textwidth}
            \centering
            \gridimg{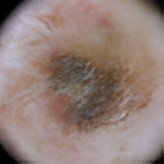}
            \parbox{\textwidth}{\centering\vspace{0.3em} Seborrheic \mbox{Keratosis}}
        \end{subfigure}
    \end{subfigure}

    \caption{Samples from ISIC (top) and their generated counterparts (bottom).}
    \label{fig:gen_quali}
\end{figure*}

\subsection{Experimental Data Setup} \label{sub:experiments:dataset}
For this study, we created two distinct datasets from the ISIC Archive~\cite{isicarchive}, a large, publicly available repository of skin lesion images. We focused on five pathological classes: \textit{actinic keratosis}, \textit{basal cell carcinoma}, \textit{melanoma}, \textit{nevus}, and \textit{seborrheic keratosis}. 
Fig.~\ref{fig:gen_quali} shows some qualitative examples for each class, as well as an example of their synthetically generated counterparts. 
Using the available metadata, images can be further categorized based on the type of acquisition device: \textit{contact polarized} (the dermatoscope touches the skin and uses polarized light), \textit{contact non-polarized} (contact without polarization), and \textit{non-contact polarized} (polarized light without skin contact). This stratification yielded two separate datasets, corresponding to the presence or absence of dermoscopic imaging acquisition device information. 
We refer to these two datasets as \textit{typed dermoscopic dataset} (that will be used for the clients' private data $D_{\text{priv}}$)  and \textit{untyped dermoscopic dataset} (used for the public data $D_{\text{pub}}$), respectively.

\begin{table}[b]
\centering
\caption{Distribution of images per pathological class.} %
\label{tab:distribution} %
\setlength{\tabcolsep}{5pt}
\renewcommand{\arraystretch}{1.1}
\resizebox{\textwidth}{!}{%
\begin{tabular}{c|c|ccccc|c}
\multirow{2}{*}{Dataset} & \multirow{2}{*}{Dermatoscope} & Actinic & Basal Cell & \multirow{2}{*}{Melanoma} & \multirow{2}{*}{Nevus} & Seborrheic & \multirow{2}{*}{Total} \\
& & Keratosis & Carcinoma & & & Keratosis & \\
\hline
typed & non-contact polarized & 54  & 208 & 194 & 270 & 64 & 790 \\
typed & contact non-polarized & 222 & 549 & 591 & 1564 & 438 & 3364\\
typed & contact polarized & 133 & 469 & 609 & 8227 & 177 & 9615 \\
\hdashline
untyped & N/A & 1119 & 2593 & 2732 & 5380 & 847 & 12671 \\
\end{tabular}
}
\end{table}

Since the same lesion may appear multiple times in the dataset %
under varying conditions -- such as different zoom levels, lighting, or acquisition settings -- we retained only one representative image per unique lesion. When multiple images of the same lesion were available, a single image was randomly selected to avoid redundancy and reduce potential bias in the dataset.

\begin{figure*}[t]
    \centering
    \begin{subfigure}{\textwidth}
        \centering
        \begin{subfigure}[t]{.28\textwidth}
            \centering
            \gridimg{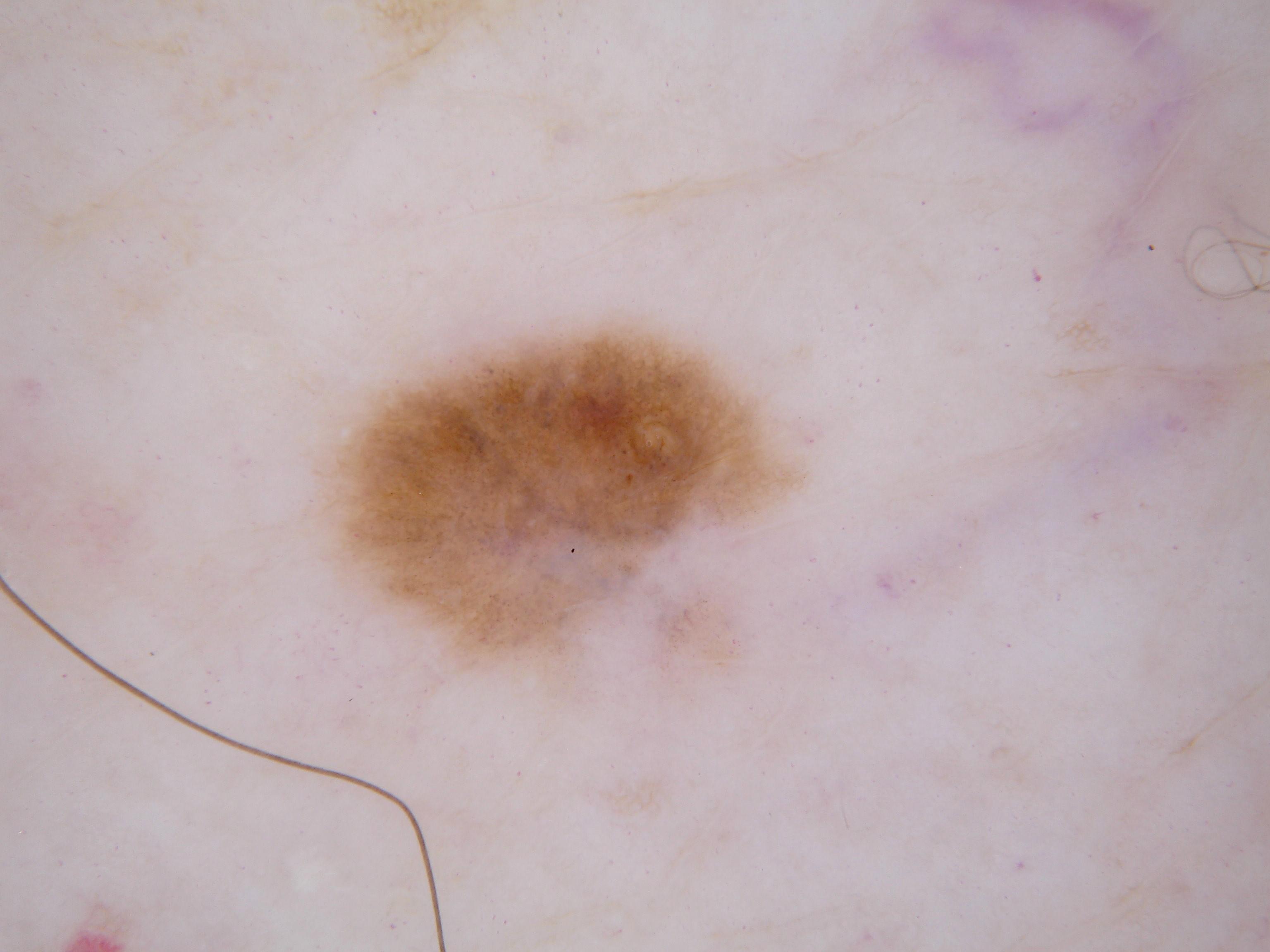}
            \parbox{\textwidth}{\centering\vspace{0.3em}  Contact Non-Polarized}
        \end{subfigure}
        \hspace{0.05\textwidth}
        \begin{subfigure}[t]{.28\textwidth}
            \centering
            \gridimg{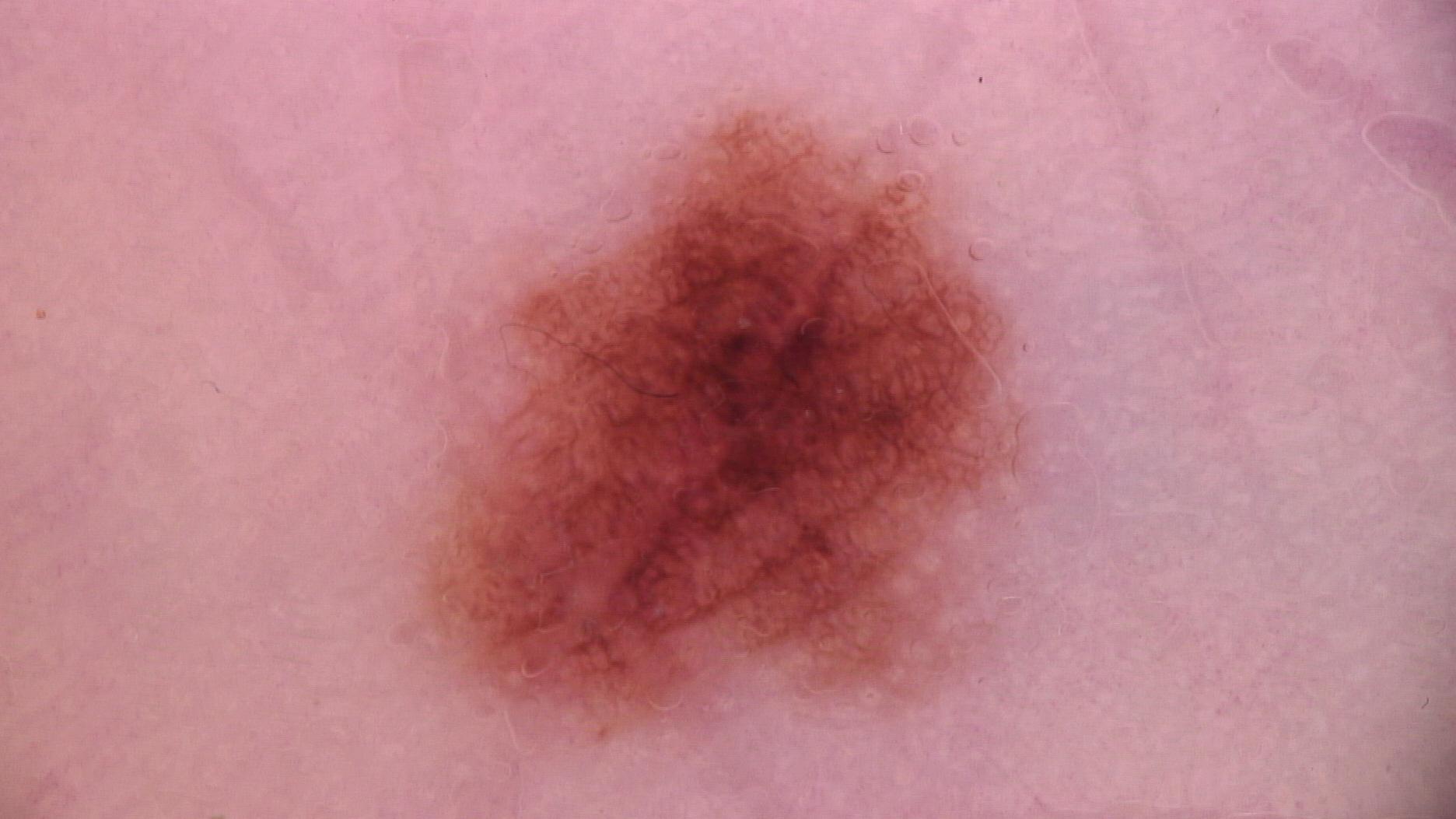}
            \parbox{\textwidth}{\centering\vspace{0.3em} Contact Polarized}
        \end{subfigure}
        \hspace{0.05\textwidth}
        \begin{subfigure}[t]{.28\textwidth}
            \centering
            \gridimg{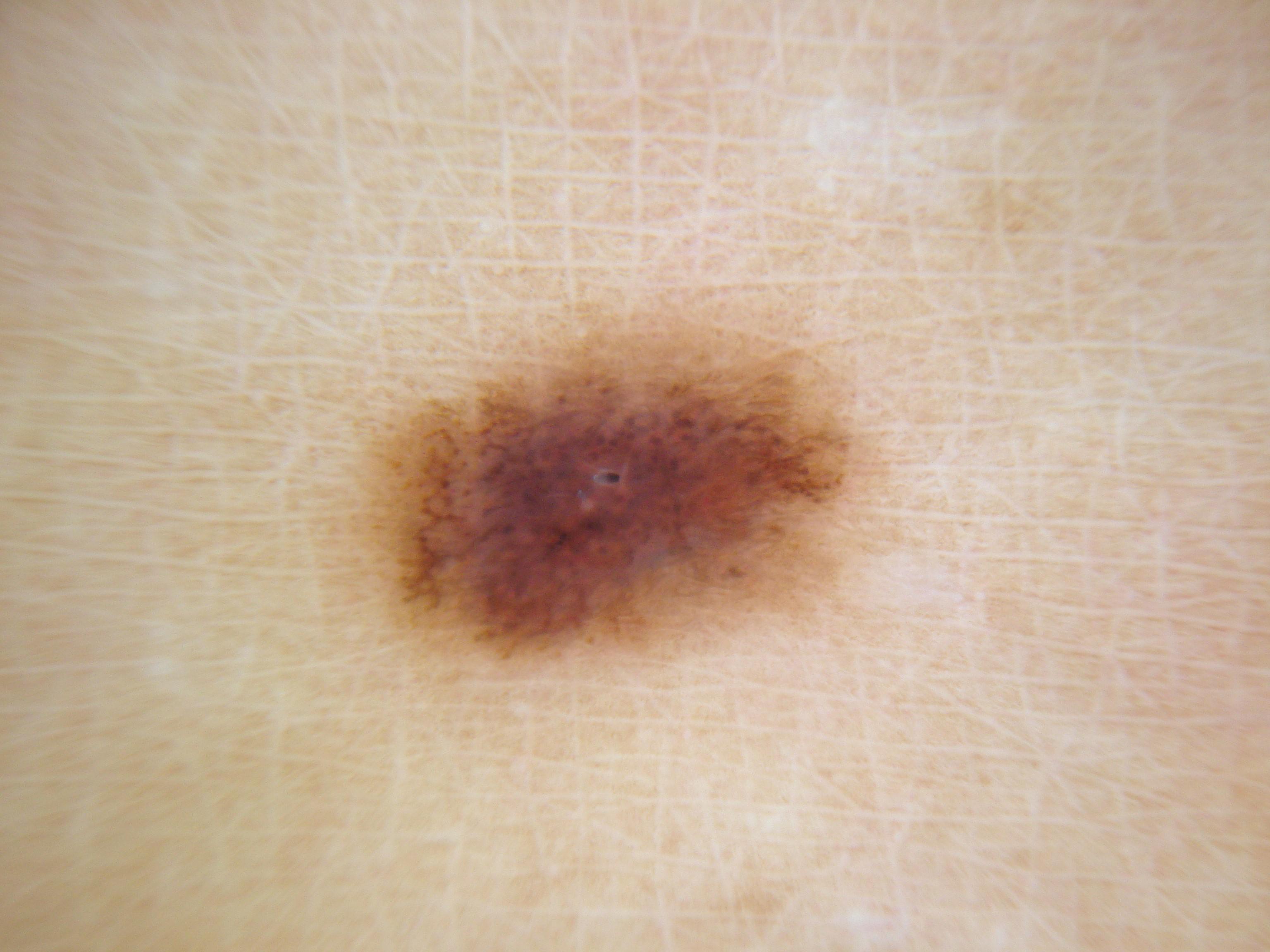}
            \parbox{\textwidth}{\centering\vspace{0.3em} Non-Contact Polarized}
        \end{subfigure}
    \end{subfigure}

    \caption{Examples of the \textit{nevus} class acquired using different imaging devices.}
    \label{fig:sensor_quali}
\end{figure*}

The \textit{untyped dermoscopic dataset} comprises a total of 12,671 images distributed across the five diagnostic categories. As summarized in Table~\ref{tab:distribution}, the dataset exhibits a notable class imbalance with the largest class %
(\textit{nevus}) having more than 5,000 samples, and the smallest
(\textit{seborrheic keratosis}) less than 1,000. %
This distribution reflects the natural prevalence of lesions in clinical practice: benign nevi are far more common than malignant or precancerous conditions, yet the rarer categories are often the most critical for accurate diagnosis~\cite{tschandl2018ham10000,combalia2019bcn20000}.
A similar pattern of imbalance is observed in the \textit{typed dermoscopic dataset}. In addition to the unequal class distribution, the number of samples also varies substantially across the three acquisition types -- \textit{contact polarized}, \textit{contact non-polarized}, and \textit{non-contact polarized} (see Fig. \ref{fig:sensor_quali} for a visual example of the differences between the sensors).
As shown in Table~\ref{tab:distribution}, the majority of images belong to the \textit{contact polarized} category, which contains 9,615 samples, predominantly \textit{nevi} (8,227 images). The \textit{contact non-polarized} subset includes 3,364 images, with a relatively higher proportion of \textit{seborrheic keratoses} (438) compared to the other subsets. Finally, the \textit{non-contact polarized} subset is the smallest, with only 790 samples in total, making it particularly underrepresented.

\subsubsection{Data Usage in Experiments} \label{sub:experiments:data}
To realistically emulate the data governance constraints typical of clinical applications, the two datasets were used in complementary roles: the \textit{typed dermoscopic dataset}, which includes device-type metadata, was treated as a private dataset, representative of hospital-held collections that cannot be centrally shared due to privacy and regulatory constraints (\eg, the GDPR). This dataset was therefore employed for %
federated learning experiments, where each client has data from a single acquisition type, allowing also for device-bias analysis. %
In contrast, the \textit{untyped dermoscopic dataset} was treated as a publicly available dataset, analogous to open-access dermatology repositories that lack detailed acquisition metadata. It was used to pretrain the EfficientNet-B0 model at server side %
and to train the U-Net–based generative network that produces the synthetic lesion images. %
This separation between \textit{private typed} and \textit{public untyped} data reflects a realistic medical setting in which sensitive clinical data must remain local while non-sensitive public data can be used to improve global model performance.

\subsubsection{Federated Splits}
To simulate a realistic federated learning scenario, training and validation data were distributed across 85 clients according to the dermoscopic acquisition device. 
Images were first grouped by device type -- \textit{contact polarized}, \textit{contact non-polarized}, and \textit{non-contact polarized} -- with the number of clients per device proportional to its data size (56, 24, and 5 clients, respectively). 
Within each device group, data were distributed across clients using a Dirichlet partition over class labels to induce controlled statistical heterogeneity. Each client is associated with a single device, capturing both class imbalance and domain heterogeneity across the federation.  
Training and validation splits followed the same strategy, while the test set was kept centralized and untouched. %

\section{Experimental Results}
\subsection{Performance Evaluation} \label{sub:experiments:main_results}

We evaluate the proposed approach against the standard federated averaging baseline (FedAvg~\cite{mcmahan2017communication}) and two representative advanced federated optimization strategies, namely MOON~\cite{li2021model} and FedProx~\cite{li2020federated}. 
To ensure a fair comparison, all methods share the same network architecture and, unless otherwise specified, are initialized from the public-data pretraining described in Section~\ref{sub:method:pretrain}. 
In addition, we report results obtained without pretraining to quantify the impact of this initialization step. 
Classification accuracy is reported separately for each acquisition device—contact polarized (CP), contact non-polarized (CNP), and non-contact polarized (NCP)—as well as averaged across devices.

\begin{table}[h]
\centering
\caption{Accuracy on the test data (\%). The Table shows the accuracy on each acquisition device (CP, CNP, NCP) and the average (Avg) value for different federated learning strategies. If pretraining is not used, we start from standard ImageNet initialization without public-data pretraining. %
}

\label{tab:res}
\setlength{\tabcolsep}{4pt}
\begin{tabular}{cc|ccc|c|ccc|c}
\multirow{2}{*}{Method} & \multirow{2}{*}{Pretraining} 
& \multicolumn{4}{c|}{Accuracy (\%)} 
& \multicolumn{4}{c}{F1-score (\%)} \\
& & CP & CNP & NCP & Avg & CP & CNP & NCP & Avg \\
\hline
FedAvg & \xmark & 91.7 & 63.2 & 75.0 & 76.6 & 62.4 & 51.6 & 62.4 & 58.8\\
MOON & \xmark & 92.2 & 66.1 & 71.3 & 76.5 & 62.8 & 54.4 & 57.0 & 58.1\\
FedPROX & \xmark & 89.4 & 66.2 & 71.3 & 75.6 & 56 & 56.8 & 52.8 & 55.2 \\
\hdashline
FedSSG & \xmark & 92.5 & 65.3 & 73.8 & 77.2 & \textbf{67.0} & 52.8 & 52.8 & 57.5\\
\hline
FedAvg & \cmark & 92.9 & 69.7 & 76.2 & 79.6 & 65.2 & 60.0 & 61.0 & 62.1\\
MOON & \cmark & 92.3 & 64.7 & 71.3 & 76.1 & 62.4 & 53.2 & 59.8 & 58.5\\
FedPROX & \cmark & 91.3 & 66.5 & 72.5 & 76.8 & 58.4 & 55.8 & 60.6 & 58.3\\
\hdashline
FedSSG & \cmark & \textbf{93.4} & \textbf{71.2} & \textbf{82.5} & \textbf{82.4} & 65.0 & \textbf{62.8} & \textbf{66.2} & \textbf{64.7} \\
\end{tabular}
\end{table}

The results are summarized in Table~\ref{tab:res}. 
When initialized from the public data pretraining, FedAvg achieves an average accuracy of $79.6\%$, while the F1-score is $62.1\%$. 
However, its performance varies substantially across devices, with high accuracy on the larger CP domain ($92.9\%$ accuracy, $65.2\%$ F1) and markedly lower results on the underrepresented CNP ($69.7\%$ accuracy, $60.0\%$ F1) and NCP ($76.2\%$ accuracy, $61.0\%$ F1) domains. 
Training FedAvg from scratch leads to a further drop of approximately $3\%$ in average accuracy and $3.3\%$ in F1, highlighting the importance of public-data pretraining in this heterogeneous federated setting.
MOON does not provide improvements over FedAvg in this scenario. 
Although it maintains competitive performance on the CP domain, it struggles on the less represented devices, resulting in a lower average accuracy. Furthermore, unlike the other approaches, it is not able to gain any advantage from the pretraining step.
FedProx yields slightly more balanced results across devices and improves marginally over MOON when pretraining is also used, but still fails to effectively address the strong domain and class imbalance, achieving an average accuracy of $76.8\%$ and $58.3\%$ F1 starting from the pretrained model.

Our generative-enhanced federated approach (FedSSG) consistently outperforms all baselines. 
With public-data pretraining, it reaches an average accuracy of $82.4\%$ ($64.7\%$ F1), corresponding to an improvement of nearly $3\%$ ($2.6\%$) over pretrained FedAvg. 
Notably, the gains are most pronounced in the underrepresented domains: while the improvement in the CP domain is modest ($+0.5\%$), accuracy increases by $+1.5\%$ on CNP and by $+6.3\%$ on NCP compared to FedAvg. The F1 scores follow a similar trend, with the only exception being non-pretrained CP, where device imbalance leads to inflated results for competitors that overfit this domain at the price of much lower results on the others.
Our approach is also better in exploiting the knowledge acquired in the pretraining step (gaining more than $5\%$ accuracy when adding this step) than competitors. 
These results indicate that the proposed generative data augmentation strategy effectively mitigates domain imbalance, stabilizes federated training, and significantly enhances generalization to less represented devices.

\subsection{Experiments in Different Federated Settings}\label{sub:experiments:fed_config}
\subsubsection{Total number of clients}
We further investigate the influence of the total number of clients participating in the federated learning process. In the main results, we set the total number of clients to 85, while here we also evaluate alternative configurations with different client populations to assess the robustness of this choice. In particular, we consider setups with 70 and 100 total clients, maintaining the same data distribution strategy and training protocol previously adopted. This analysis allows us to quantify the effect of the client population size on overall performance and to verify whether the proposed configuration represents a suitable trade-off between scalability and accuracy. The quantitative results are reported in Table~\ref{tab:tot_clients}, which shows that performances are stable with some small variations across the different tested numbers of clients. Our approach consistently outperforms the FedAVG baseline in all settings.

\begin{table}[tbp]
    \centering
    \caption{Performances varying the total number of clients.}
    \label{tab:tot_clients}
    \setlength{\tabcolsep}{4pt}
    \begin{tabular}{c|c|ccc:c|ccc:c}
      \multirow{2}{*}{Total clients} 
      & \multirow{2}{*}{Approach}
      & \multicolumn{4}{c|}{Accuracy (\%)} 
      & \multicolumn{4}{c}{F1-score (\%)} \\
      & & CP & CNP & NCP & Avg & CP & CNP & NCP & Avg \\
      \hline
      \multirow{2}{*}{70} 
      & FedAVG & \textbf{92.9} & 67.1 & \textbf{78.8} & 79.6 & 62.4 & 54.4 & 62.4 & 59.7\\
      & FedSSG & 92.5 & \textbf{68.8} & \textbf{78.8} & \textbf{80.0} & \textbf{63.6} & \textbf{59.2} & \textbf{63.0} & \textbf{61.9} \\
      \hdashline
      \multirow{2}{*}{\underline{85}}
      & FedAVG & 92.9 & 69.7 & 76.2 & 79.6 & \textbf{65.2} & 60.0 & 61.0 & 62.1\\
      & FedSSG & \textbf{93.4} & \textbf{71.2} & \textbf{82.5} & \textbf{82.4} & 65.0 & \textbf{62.8} & \textbf{66.2} & \textbf{64.7} \\
      \hdashline
      \multirow{2}{*}{100}
      & FedAVG & 92.7 & \textbf{67.9} & 76.3 & 79.0 & 63.8 & 55.4 & 63.6 & 60.9\\ 
      & FedSSG & \textbf{93.1} & 66.1  & \textbf{82.5}  & \textbf{80.6} & \textbf{65.6} & \textbf{55.8} & \textbf{76.2} & \textbf{64.7} \\
    \end{tabular}
\end{table}

\subsubsection{Number of active clients per round}
We conducted an ablation study to analyze the impact of the number of active clients per communication round. In the default setting, 6 clients are randomly selected and activated at each round. To assess the effect of this design choice, we performed additional experiments by varying the number of active clients per round, considering configurations with 4 and 8 active clients, while keeping all other training parameters unchanged. The results of this comparison are reported in Table~\ref{tab:active_clients} and show stable gains of FedSSG over the FedAVG reference across the different settings.

\begin{table}[tbp]
    \centering
    \caption{Ablation study on the number of active clients per round.}
    \label{tab:active_clients}
    \setlength{\tabcolsep}{4pt}
    \begin{tabular}{c|c|ccc:c|ccc:c}
      \multirow{2}{*}{Active clients} 
      & \multirow{2}{*}{Approach} 
      & \multicolumn{4}{c|}{Accuracy (\%)} 
      & \multicolumn{4}{c}{F1-score (\%)} \\
      & & CP & CNP & NCP & Avg & CP & CNP & NCP & Avg \\
      \hline
      \multirow{2}{*}{4} 
      & FedAVG & 93.0 & \textbf{71.5} & 77.5 & 80.7 & \textbf{64.0} & \textbf{60.2} & 61.2 &  61.8\\
      & FedSSG & \textbf{93.1} & 69.7 & \textbf{82.5} & \textbf{81.8} & 62.6 & 59.2 & \textbf{73.2} & \textbf{65.0} \\
      \hdashline
      \multirow{2}{*}{\underline{6}} 
      & FedAVG & 92.9 & 69.7 & 76.2 & 79.6 & \textbf{65.2} & 60.0 & 61.0 & 62.1\\
      & FedSSG & \textbf{93.4} & \textbf{71.2} & \textbf{82.5} & \textbf{82.4} & 65.0 & \textbf{62.8} & \textbf{66.2} & \textbf{64.7} \\
      \hdashline
      \multirow{2}{*}{8} 
      & FedAVG & 93.1 & 68.0 & 77.5 & 79.5 & 62.6 & 55.4 & 63.6 & 60.5 \\
      & FedSSG & \textbf{94.2} & \textbf{70.9} & \textbf{78.8} & \textbf{81.3} & \textbf{67.0} & \textbf{60.0} & \textbf{67.2} & \textbf{64.7} \\
    \end{tabular}
\end{table}

\subsection{Ablation Studies}
\label{sub:experiments:ablation}

\subsubsection{Number of Generated Images} 
As a first ablation experiment, we analyzed the results obtained by using the same fixed number of generated images for all domains, set to 50 samples per domain. We then evaluated the impact of using the strategy of Section \ref{sec:method} but modifying the overall amount of synthetic data by either doubling or halving the number of generated samples. The results are reported in Table~\ref{tab:gen_num}.
As shown in the table, using a fixed number of generated images leads to reduced performance. The same happens when simply scaling the number of generated samples by halving or doubling it. In contrast, the proposed strategy achieves better results, especially on the CNP and NCP subsets, leading to the highest average performance.

\begin{table}[tbp]
    \centering
    \caption{Ablation study on $S_{d(k)}$}
    \label{tab:gen_num}
    \setlength{\tabcolsep}{5pt}
    \renewcommand{\arraystretch}{1.2}
    \begin{tabular}{c|cccc|cccc}
      \multirow{2}{*}{$S_{d(k)}$} 
      & \multicolumn{4}{c|}{Accuracy (\%)} 
      & \multicolumn{4}{c}{F1-score (\%)} \\
      & CP & CNP & NCP & Avg & CP & CNP & NCP & Avg \\
      \hline
      $(50, 50, 50)$ & 93.7 & 69.7 & 81.2 & 81.5 & 68.0 & 58.6 & 66.6 & 64.4\\
      \hdashline
      $(10, 25, 40)$ & 93.4 & 68.3 & 80.0 & 80.6 & 63.0 & 57.6 & 64.8 & 61.8\\
      \underline{$(20, 50, 80)$} & 93.4 & 71.2 & 82.5 & 82.4 & 65.0 & 62.8 & 66.2 & 64.7\\
      $(40, 100, 160)$ & 93.6 & 69.7 & 80.0 & 81.1 & 65.4 & 59.8 & 71.0 & 65.4\\
    \end{tabular}
\end{table}

\subsubsection{Longer Training} 
We also tested whether longer training could lead to better performances. Training our federated approach for 300 epochs leads to accuracies of $93.7\%$ (CP), $71.8\%$ (CNP), and  $81.2\%$ (NCP), leading to a slightly lower average accuracy of $82.2\%$, thus showing that the approach has already converged after 150 epochs and there is no need for longer training procedures. The F1-scores closely follow the accuracy in this case as well.

\section{Conclusion}
In this work, we introduced a federated learning framework tailored to multi-domain and class-imbalanced medical image classification, and validated it on a challenging skin lesion classification benchmark. Our approach leverages publicly available data to perform server-side pretraining of both a classifier and a class-conditional diffusion model, providing a robust initialization for federated optimization under strong data heterogeneity. The pretrained generative model is then deployed to clients grouped by acquisition device and used to synthesize targeted samples that compensate for both rare pathologies and underrepresented imaging domains. By integrating generative augmentation directly into the federated training process, our method improves training stability and achieves consistent performance gains, exceeding the best competing federated baselines by almost $3\%$ in average accuracy, with particularly strong improvements on minority device domains.
Extensive ablation studies further confirm the effectiveness and robustness of the proposed framework. In future work, we aim to extend this approach to a broader range of medical imaging modalities and tasks and to further enhance its ability to generalize across unseen acquisition devices and institutional settings.

\bibliographystyle{splncs04}
\bibliography{main}

\begin{thebibliography}{10}
\providecommand{\url}[1]{\texttt{#1}}
\providecommand{\urlprefix}{URL }
\providecommand{\doi}[1]{https://doi.org/#1}

\bibitem{beutel2020flower}
Beutel, D.J., Topal, T., Mathur, A., Qiu, X., Fernandez-Marques, J., Gao, Y.,
  Sani, L., Kwing, H.L., Parcollet, T., Gusmão, P.P.d., Lane, N.D.: Flower: A
  friendly federated learning research framework. arXiv preprint
  arXiv:2007.14390  (2020)

\bibitem{info11020125}
Buslaev, A., Iglovikov, V.I., Khvedchenya, E., Parinov, A., Druzhinin, M.,
  Kalinin, A.A.: Albumentations: Fast and flexible image augmentations.
  Information  \textbf{11}(2) (2020)

\bibitem{caligiuri2025fedpromo}
Caligiuri, M., Barbato, F., Shenaj, D., Michieli, U., Zanuttigh, P.: Fedpromo:
  Federated lightweight proxy models at the edge bring new domains to
  foundation models. arXiv preprint arXiv:2508.03356  (2025)

\bibitem{chen2023on}
Chen, H.Y., Tu, C.H., Li, Z., Shen, H.W., Chao, W.L.: On the importance and
  applicability of pre-training for federated learning. In: Proceedings of the
  11th International Conference on Learning Representations (ICLR) (2023)

\bibitem{deng2009imagenet}
Deng, J., Dong, W., Socher, R., Li, L.J., Li, K., Fei-Fei, L.: Imagenet: A
  large-scale hierarchical image database. In: 2009 IEEE conference on computer
  vision and pattern recognition. pp. 248--255 (2009)

\bibitem{guan2024federated}
Guan, H., Yap, P.T., Bozoki, A., Liu, M.: Federated learning for medical image
  analysis: A survey. Pattern Recognition  \textbf{151},  110424 (2024).
  \doi{10.1016/j.patcog.2024.110424}

\bibitem{combalia2019bcn20000}
Hern\'andez-P\'erez, C., Combalia, M., Podlipnik, S., Codella, N.C., Rotemberg,
  V., Halpern, A.C., Reiter, O., Carrera, C., Barreiro, A., Helba, B., Puig,
  S., Vilaplana, V., Malvehy, J.: Bcn20000: Dermoscopic lesions in the wild.
  Scientific Data  \textbf{11}(1), ~641 (2024).
  \doi{10.1038/s41597-024-03387-w}

\bibitem{isicarchive}
{ISIC Collaboration}: {International Skin Imaging Collaboration (ISIC)
  Archive.} (Accessed: 2025-10-22), \url{https://www.isic-archive.com/}

\bibitem{kairouz2021advances}
Kairouz, P., McMahan, H.B., Avent, B., Bellet, A., Bennis, M., Bhagoji, A.N.,
  Bonawitz, K., Charles, Z., Cormode, G., Cummings, R., et~al.: Advances and
  open problems in federated learning. Foundations and Trends® in Machine
  Learning  \textbf{14}(1--2),  1--210 (2021)

\bibitem{kaissis2021secure}
Kaissis, G., Makowski, M.R., Rückert, D., Braren, R.F.: Secure,
  privacy-preserving and federated machine learning in medical imaging. Nature
  Machine Intelligence  \textbf{3}(6),  305--311 (2021)

\bibitem{kaissis2020secure}
Kaissis, G.A., Makowski, M.R., Rückert, D., Braren, R.F.: Secure,
  privacy-preserving and federated machine learning in medical imaging. Nature
  Machine Intelligence  \textbf{2}(6),  305--311 (2020)

\bibitem{kamran2025fedgan}
Kamran, H., Hussain, S.J., Latif, S., Soomro, I.A., Alnfiai, M.M., Alotaibi,
  N.N.: Fedgan: Federated diabetic retinopathy image generation. PLOS ONE
  \textbf{20}(7),  e0326579 (2025). \doi{10.1371/journal.pone.0326579}

\bibitem{karimireddy2020scaffold}
Karimireddy, S.P., Kale, S., Mohri, M., Reddi, S., Stich, S., Suresh, A.T.:
  Scaffold: Stochastic controlled averaging for federated learning. In:
  International Conference on Machine Learning. pp. 5132--5143. PMLR (2020)

\bibitem{li2021model}
Li, Q., He, B., Song, D.: Model-contrastive federated learning. In: IEEE/CVF
  Conference on Computer Vision and Pattern Recognition. pp. 10713--10722
  (2021)

\bibitem{li2024synergizing}
Li, S., Ye, F., Fang, M., Zhao, J., Chan, Y.H., Ngai, E.C.H., Voigt, T.:
  Synergizing foundation models and federated learning: A survey. arXiv
  preprint arXiv:2406.12844  (2024)

\bibitem{li2020federated}
Li, T., Sahu, A.K., Zaheer, M., Sanjabi, M., Talwalkar, A., Smith, V.:
  Federated optimization in heterogeneous networks. Proceedings of Machine
  Learning and Systems  \textbf{2},  429--450 (2020)

\bibitem{litjens2017survey}
Litjens, G., Kooi, T., Bejnordi, B.E., Setio, A.A.A., Ciompi, F., Ghafoorian,
  M., van~der Laak, J.A., van Ginneken, B., Sánchez, C.I.: A survey on deep
  learning in medical image analysis. Medical image analysis  \textbf{42},
  60--88 (2017)

\bibitem{liu2022convnet2020s}
Liu, Z., Mao, H., Wu, C.Y., Feichtenhofer, C., Darrell, T., Xie, S.: A convnet
  for the 2020s (2022), \url{https://arxiv.org/abs/2201.03545}

\bibitem{loshchilov2017decoupled}
Loshchilov, I., Hutter, F.: Decoupled weight decay regularization. In:
  International Conference on Learning Representations (ICLR) (2017)

\bibitem{loshchilov2017sgdr}
Loshchilov, I., Hutter, F.: Sgdr: Stochastic gradient descent with warm
  restarts. In: International Conference on Learning Representations (ICLR)
  (2017)

\bibitem{mcmahan2017communication}
McMahan, B., Moore, E., Ramage, D., Hampson, S., y~Arcas, B.A.:
  Communication-efficient learning of deep networks from decentralized data.
  In: Artificial Intelligence and Statistics. pp. 1273--1282. PMLR (2017)

\bibitem{nazir2023diagnostics}
Nazir, S., Kaleem, M.: Federated learning for medical image analysis with deep
  neural networks. Diagnostics  \textbf{13}(9), ~1532 (2023)

\bibitem{nguyen2023where}
Nguyen, J., Wang, J., Malik, K., Sanjabi, M., Rabbat, M.: Where to begin? on
  the impact of pre-training and initialization in federated learning. In:
  Proceedings of the 11th International Conference on Learning Representations
  (ICLR) (2023)

\bibitem{perez2017filmvisualreasoninggeneral}
Perez, E., Strub, F., de~Vries, H., Dumoulin, V., Courville, A.: Film: Visual
  reasoning with a general conditioning layer (2017),
  \url{https://arxiv.org/abs/1709.07871}

\bibitem{sheller2020federated}
Sheller, M.J., Edwards, B., Reina, D.G., Martin, J., Pati, S., Kotrotsou, A.,
  Milchenko, M., Xu, W., Marcus, D., Colen, R.R., Bakas, S.: Federated learning
  in medicine: facilitating multi-institutional collaborations without sharing
  patient data. Scientific reports  \textbf{10}(1),  1--12 (2020)

\bibitem{shen2017deep}
Shen, D., Wu, G., Suk, H.I.: Deep learning in medical image analysis. Annual
  review of biomedical engineering  \textbf{19},  221--248 (2017)

\bibitem{shenaj2023learning}
Shenaj, D., Fan{\`\i}, E., Toldo, M., Caldarola, D., Tavera, A., Michieli, U.,
  Ciccone, M., Zanuttigh, P., Caputo, B.: Learning across domains and devices:
  Style-driven source-free domain adaptation in clustered federated learning.
  In: IEEE/CVF Winter Conference on Applications of Computer Vision. pp.
  444--454 (2023)

\bibitem{shenaj2023federated}
Shenaj, D., Rizzoli, G., Zanuttigh, P.: Federated learning in computer vision.
  IEEE Access  (2023)

\bibitem{tan2019efficientnet}
Tan, M., Le, Q.: Efficientnet: Rethinking model scaling for convolutional
  neural networks. In: International Conference on Machine Learning. PMLR
  (2019)

\bibitem{tschandl2018ham10000}
Tschandl, P., Rosendahl, C., Kittler, H.: The ham10000 dataset, a large
  collection of multi-source dermatoscopic images of common pigmented skin
  lesions. Scientific Data  \textbf{5}(1),  180161 (2018).
  \doi{10.1038/sdata.2018.161}

\bibitem{wang2021realesrgan}
Wang, X., Yu, R., Wu, J., Gu, K., Liu, C., Dong, W., Loy, C.C., Qiao, Y.:
  Real-esrgan: Training real-world blind super-resolution with pure synthetic
  data. In: Proceedings of the IEEE/CVF International Conference on Computer
  Vision (ICCV). pp. 1905--1914 (2021)

\bibitem{wu2023fediic}
Wu, N., Yu, L., Yang, X., Cheng, K.T., Yan, Z.: Fediic: Towards robust
  federated learning for class-imbalanced medical image classification. In:
  Medical Image Computing and Computer Assisted Intervention – MICCAI. pp.
  692--702. Springer (2023). \doi{10.1007/978-3-031-43895-0_65}

\bibitem{zhou2021fed}
Zhou, Y., Yao, Z., Xu, X., Yang, Y.: Fedbn: Federated learning on non-iid
  features via local batch normalization. arXiv preprint arXiv:2106.00645
  (2021)

\bibitem{zhou2025flmedical}
Zhou, Z., Luo, G., Chen, M., Weng, Z., Zhu, Y.: Federated learning for medical
  image classification: A comprehensive benchmark. arXiv preprint  (2025)

\bibitem{zhuang2025foundationmodelmeetsfederated}
Zhuang, W., Chen, C., Li, J., Chen, C., Jin, Y., Lyu, L.: When foundation model
  meets federated learning: Motivations, challenges, and future directions.
  arXiv  (2025)

\end{thebibliography}

\end{document}